\documentclass[letterpaper, 10 pt, conference]{ieeeconf}  
\IEEEoverridecommandlockouts                              

\usepackage{00_ps}

\preprinttrue
\publishedfalse

\glsdisablehyper 
\DeclareAcronym{clf}{
    short = CLF,
    long  = Control Lyapunov Function,
}
\DeclareAcronym{cav}{
    short = CAV,
    long  = Connected and Automated Vehicle,
}
\DeclareAcronym{cbf}{
    short = CBF,
    long  = Control Barrier Function,
}
\DeclareAcronym{cg}{
    short = CG,
    long = Center of Gravity,
    short-plural = s,
    long-plural-form = Centers of Gravity,
}
\DeclareAcronym{cnn}{
    short = CNN,
    long  = Convolutional Neural Network
}
\DeclareAcronym{cpm}{
    short = CPM,
    long  = Cyber-Physical Mobility
}
\DeclareAcronym{cpmlab}{
    short = CPM Lab,
    long  = \ac{cpm} Lab
}
\DeclareAcronym{dmpc}{
    short = DMPC,
    long  = distributed model predictive control
}
\DeclareAcronym{dql}{
    short = DQL,
    long  = Deep Q-Learning
}

\DeclareAcronym{il}{
    short = IL,
    long  = Imitation Learning,
    short-indefinite = an,
}
\DeclareAcronym{mappo}{
    short = MAPPO,
    long  = Multi-Agent \ac{ppo},
    short-indefinite = an,
}
\DeclareAcronym{maddpg}{
    short = MADDPG,
    long  = Multi-Agent Deep Deterministic Policy Gradient,
    short-indefinite = an,
}
\DeclareAcronym{mas}{
    short = MAS,
    long  = Multi-Agent System,
    short-indefinite = an,
}
\DeclareAcronym{mdp}{
    short = MDP,
    long  = Markov decision process,
    short-indefinite = an,
}
\DeclareAcronym{mg}{
    short = MG,
    long  = Markov Game,
    short-indefinite = an,
}
\DeclareAcronym{mpc}{
    short = MPC,
    long  = Model Predictive Control,
    short-indefinite = an,
}
\DeclareAcronym{marl}{
    short = MARL,
    long  = Multi-Agent Reinforcement Learning,
    short-indefinite = an,
}
\DeclareAcronym{ocp}{
    short = OCP,
    long  = Optimal Control Problem,
    short-indefinite = an,
    long-indefinite = an,
}
\DeclareAcronym{per}{
    short = PER,
    long  = Prioritized Experience Replay
}
\DeclareAcronym{pomdp}{
    short = POMDP,
    long  = Partially Observable \ac{mdp}
}
\DeclareAcronym{pomg}{
    short = POMG,
    long  = Partially Observable \ac{mg}
}
\DeclareAcronym{ppo}{
    short = PPO,
    long  = Proximal Policy Optimization
}
\DeclareAcronym{qp}{
    short = QP,
    long  = Quadratic Program,
}
\DeclareAcronym{rhc}{
    short = RHC,
    long  = receding horizon control,
    short-indefinite = an,
}
\DeclareAcronym{rl}{
    short = RL,
    long  = Reinforcement Learning,
    short-indefinite = a,
}
\DeclareAcronym{ttcbf}{
    short = TTCBF,
    long  = Truncated Taylor \ac{CBF},
}
\DeclareAcronym{ttc}{
    short = TTC,
    long  = time-to-collision,
}
\DeclareAcronym{zsg}{
    short = ZSG,
    long  = Zero-Shot Generalization,
}
\providecommand{\doi}{????/placeholder-doi} 

\makeatletter
\def\ps@IEEEtitlepagestyle{%
    \def\@oddfoot{\mycopyrightnotice}%
    \def\@evenfoot{}%
}
\makeatother

\ifpreprint
    \ifpublished
        \def\mycopyrightnotice{%
            \begin{minipage}{\textwidth}
                \centering \scriptsize
                \href{https://doi.org/\doi}{DOI: \doi} \copyright \the\year{} \copyrightOwnerFull{}. 
                Personal use of this material is permitted. Permission from \copyrightOwnerShort{} must be obtained for all other uses, in any current or future media, including reprinting/republishing this material for advertising or promotional purposes, creating new collective works, for resale or redistribution to servers or lists, or reuse of any copyrighted component of this work in other works.\hfill
            \end{minipage}
            \gdef\mycopyrightnotice{}
        }
    \else 
        \def\mycopyrightnotice{%
            \begin{minipage}{\textwidth}
                \centering \scriptsize
                This paper has been accepted for publication in the Proceedings of the 2026 IEEE International Conference on Intelligent Transportation Systems (ITSC 2026). Copyright may be transferred without notice, after which this version may no longer be accessible.
            \end{minipage}
            \gdef\mycopyrightnotice{}
        }
    \fi
\else
    \def\mycopyrightnotice{}
\fi

\def\BibTeX{{\rm B\kern-.05em{\sc i\kern-.025em b}\kern-.08em
  T\kern-.1667em\lower.7ex\hbox{E}\kern-.125emX}}

\begin{document}
\title{\LARGE \bf
    Beyond Safety Filtering: Control Barrier Function-Informed Reinforcement Learning for Connected and Automated Vehicles
    \thanks{This research was supported by the Bundesministerium für Digitales und Verkehr (German Federal Ministry for Digital and Transport) within the project ``Harmonizing Mobility'' (grant number 19FS2035A).}
}

\author{
    Jianye Xu$^{1}$\,\orcidlink{0009-0001-0150-2147},~\IEEEmembership{Student~Member,~IEEE},
    Bassam Alrifaee$^{2}$\,\orcidlink{0000-0002-5982-021X},~\IEEEmembership{Senior Member, ~IEEE}
    \thanks{$^{1}$Department of Computer Science, RWTH Aachen University, Germany, \texttt{xu@embedded.rwth-aachen.de}}
    \thanks{$^{2}$Department of Aerospace Engineering, University of the Bundeswehr Munich, Germany, \texttt{bassam.alrifaee@unibw.de}}
}
    \maketitle
\thispagestyle{IEEEtitlepagestyle}
\begin{abstract}
\noindent
Reinforcement Learning (RL) uses rewards to guide learning, yet reward design is typically hand-crafted using heuristics that can be difficult to tune. We propose a Control Barrier Function (CBF)-informed reward design for Multi-Agent RL (MARL) that converts CBF constraint values under joint MARL actions into a reward signal that explicitly guides safe learning. We compare against two heuristic reward baselines in a four-way multi-lane intersection with connected and automated vehicles. Results show that our method achieves the highest task performance and is less sensitive to reward hyperparameters, yielding consistently strong performance across the tested hyperparameter range. Code for reproducing the experimental results and a video demonstration are available at \href{https://github.com/bassamlab/SigmaRL}{https://github.com/bassamlab/SigmaRL}.
\par\medskip
\end{abstract}

\noindent

\acresetall
\section{Introduction}\label{sec:introduction}
\ac{rl} is widely used for decision-making in autonomous driving, where an agent learns a control policy by maximizing a cumulative reward through interaction with an environment \cite{kiran2022deep}. A reward function in \ac{rl} encodes task objectives such as goal reaching and safety, directly shaping the learned behavior.

Reward design is central to the performance and safety of \ac{rl}, yet it often remains heuristic. Reward misspecification and unintended incentives, which can lead to unsafe behaviors, have been analyzed in depth for autonomous driving settings \cite{knox2023reward}. A range of reward formulations have been used in driving tasks, including sparse rewards for success or collisions \cite{chen2019modelfree,chen2022interpretable}, distance-based reward shaping for safe distances \cite{wang2017formulation,wu2023uncertaintyaware}, and \ac{ttc}-based reward shaping to penalize imminent collisions based on a constant-velocity assumption \cite{zhu2020safe}. These methods can work well empirically, but they generally do not connect the reward directly to a formal safety condition and often require extensive tuning across scenarios.

\acp{cbf} provide a mathematically grounded way to enforce safety by certifying forward invariance of a prescribed safe set under closed-loop dynamics \cite{ames2019control}. A standard approach is to formulate a \ac{qp} with \ac{cbf} constraints that acts as a posterior safety filter to certify nominal control actions and, if necessary, minimally modifies them to satisfy safety constraints online. Such safety filtering connects to earlier safety architectures such as Simplex \cite{seto1998simplex} and relates to barrier-based safety verification \cite{prajna2007framework}. Methods like predictive safety filtering also share the same high-level idea \cite{wabersich2021predictive}. More recently, \acp{cbf} have also been integrated into \ac{rl} to ensure safety during learning, for example, for an inverted pendulum \cite{cheng2019endtoend}, a Segway system \cite{taylor2020learning}, locomotion \cite{csomay-shanklin2021episodic}, and autonomous driving \cite{marvi2021safe,gangopadhyay2023safe,zhang2025control,xu2025learningbased}.

Despite their strong safety guarantees, online filtering can be computationally difficult to scale when safety constraints are dense and coupled, as in multi-vehicle interactions at an intersection. In this work, instead of using \acp{cbf} as a posterior safety filter, we convert the \ac{cbf} constraint satisfaction degree into an \ac{rl} reward signal. This design directly shapes the learning objective of \ac{rl} toward actions that satisfy safety constraints, aiming to reduce unsafe behaviors at the policy level. Although prior work proposes similar ideas, it is primarily limited to locomotion \cite{nilaksh2024barrier,kim2024not,kim2025learning}. Our work stands out by proposing \ac{cbf} constructions that consider coupled safety constraints in multi-vehicle interactions.

Our contributions are twofold: 1) We propose a \ac{cbf}-informed reward design for \ac{marl} in multi-vehicle settings, where reward signals are computed from \ac{cbf} constraint values under joint \ac{marl} actions to explicitly guide the learning process toward safety; 2) We evaluate the proposed method in a four-way multi-lane intersection and compare against two representative heuristic reward baselines (distance-based \cite{wu2023uncertaintyaware} and \ac{ttc}-based \cite{zhu2020safe}). Results show that our method achieves the best overall task performance and is more robust to reward hyperparameter choices, with reduced sensitivity across tested hyperparameter settings.

The remainder of the paper is as follows.
\cref{sec:problem-statement} formally formulates the multi-vehicle intersection problem and evaluation metrics. \cref{sec:cbf-informed-marl} presents our \ac{cbf} construction and \ac{cbf}-informed reward design. \cref{sec:eva} reports the experimental setup and quantitative results. \cref{sec:conclusions} concludes the paper.

\section{Problem Statement} \label{sec:problem-statement}
\begin{figure}[t!]
    \centering
    \includegraphics[width=0.90\linewidth]{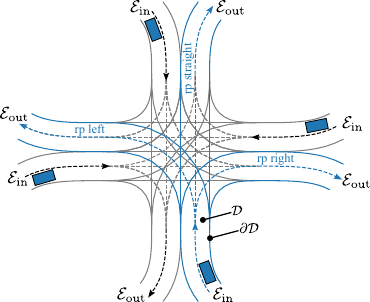}
    \caption{Overview of a four-way multi-lane intersection. $\mathcal{E}_\mathrm{in}$: entry region; $\mathcal{E}_\mathrm{out}$: exit region; $\mathrm{rp}$: reference path; $\mathcal{D}$: drivable corridor; $\partial \mathcal{D}$: road boundary.}
    \label{fig_intersection}
\end{figure}

We consider a four-way multi-lane intersection in a planar workspace $\mathcal{W}\subset\mathbb{R}^2$ with four entry regions and four exit regions, as depicted in \cref{fig_intersection}. A set of $N$ \acp{cav}, indexed by $\mathcal{I}=\{1,\dots,N\}$, enter the intersection from the entry regions and target their respective exit regions. Each vehicle $i\in\mathcal{I}$ is assigned a reference path
$\mathrm{rp}^i:[0,1]\to\mathcal{W}$,
where the normalized path coordinate $s\in[0,1]$ satisfies $\mathrm{rp}^i(0)\in\mathcal{E}_{\mathrm{in}}^i\subset \mathcal{W}$ and $\mathrm{rp}^i(1)\in\mathcal{E}_{\mathrm{out}}^i\subset \mathcal{W}$. Here, $\mathcal{E}_{\mathrm{in}}^i$ and $\mathcal{E}_{\mathrm{out}}^i$ denote the entry and exit regions of vehicle $i$, respectively. 
For each entry region, three candidate reference paths are available, corresponding to a left turn, going straight, and a right turn (see the blue annotations in \cref{fig_intersection}). Associated with $\mathrm{rp}^i$ is a drivable corridor $\mathcal{D}^i\subset\mathcal{W}$, bounded by the left road boundary $\partial\mathcal{D}^{i,\mathrm{L}}$ and the right road boundary $\partial\mathcal{D}^{i,\mathrm{R}}$. Let $\partial\mathcal{D}^{i} \coloneqq \partial\mathcal{D}^{i,\mathrm{L}} \cup \partial\mathcal{D}^{i,\mathrm{R}}  \subset \mathcal{W}$ denote the union of the road boundaries of vehicle $i$.

For each $i\in\mathcal{I}$, the state and control input are denoted by $\bm{x}^i\in\mathcal{X}^i\subset\mathbb{R}^n$ and $\bm{u}^i\in\mathcal{U}^i\subset\mathbb{R}^m$, where $\mathcal{X}^i$ denotes the state space, $\mathcal{U}^i$ the control input space, $n$ the state dimension, and $m$ the control dimension. Denote the position of vehicle $i$ by $p^i\in\mathcal{W}$ as a component of $\bm{x}^i$. Each vehicle $i$ is initialized with $p^i_0\in\mathcal{E}_{\mathrm{in}}^i$ and an initial speed sampled from $[0,v^i_{\max}]$, where $v^i_{\max}>0$ denotes its maximum admissible speed. We consider a discrete-time domain with sampling period $\Delta t>0$. The sampled state and input are denoted by
$\bm{x}^i_k=\bm{x}^i(k\Delta t), \bm{u}^i_k=\bm{u}^i(k\Delta t)$ for time step $k\in\mathbb{N}$.

Let $\mathcal{O}^i(\bm{x}^i)\subset\mathcal{W}$ denote the occupied set (footprint) of vehicle $i$ at state $\bm{x}^i$. At time step $k$, a vehicle-road collision for vehicle $i$ occurs if
\begin{equation} \label{eq:road_collision}
\mathcal{O}^i(\bm{x}^i_k)\cap\partial\mathcal{D}^i\neq\emptyset
\end{equation}
and an inter-vehicle collision between $i$ and $j$ occurs if
\begin{equation} \label{eq:vehicle_collision}
\mathcal{O}^i(\bm{x}^i_k)\cap\mathcal{O}^j(\bm{x}^j_k)\neq\emptyset.
\end{equation}
If a vehicle-road collision occurs, the involved vehicle is randomly reset to one of the four entry regions with a newly sampled initial speed. Upon reset, the vehicle is assigned a new random reference path $\mathrm{rp}^i$. If an inter-vehicle collision occurs, both involved vehicles are similarly reset.

Let $T_{\mathrm{eval}}>0$ denote a fixed evaluation time. We evaluate a policy using an event-based total reward with an additional driving-comfort penalty over all vehicles on $[0,T_{\mathrm{eval}}]$. Specifically, each time a vehicle successfully reaches its exit region, a reward of $+1$ is issued; each time a vehicle collides with a road boundary or another vehicle, a reward of $-1$ is issued. Let $N_{\mathrm{exit}}(T_{\mathrm{eval}})$ denote the total number of exit events, and let $N_{\mathrm{col}}(T_{\mathrm{eval}})$ denote the total number of collision events. In addition, we penalize uncomfortable driving behaviors using longitudinal acceleration $a^i_k$ and jerk $j^i_k$ \cite{nilsson2017lane}. 
Let $K_{\mathrm{eval}}\coloneqq \lfloor T_{\mathrm{eval}}/\Delta t\rfloor$ denote the total number of evaluation time steps. The total reward is defined as
\begin{align} \label{eq:prob-total-reward}
&R_{\mathrm{tot}}(T_{\mathrm{eval}})
=
N_{\mathrm{exit}}(T_{\mathrm{eval}})
-
N_{\mathrm{col}}(T_{\mathrm{eval}})
\nonumber\\
&\quad
-
\frac{w_{\mathrm{comf}}}{N K_{\mathrm{eval}}}
\sum_{i\in\mathcal{I}}
\sum_{k=1}^{K_{\mathrm{eval}}}
\Big(
\Big(\frac{|a^i_k|}{a_{\mathrm{norm}}}\Big)^2
+
\Big(\frac{|j^i_k|}{j_{\mathrm{norm}}}\Big)^2
\Big),
\end{align}
where $w_{\mathrm{comf}}>0$ is a weighting parameter balancing event-based rewards and uncomfortable-driving penalties, and $a_{\mathrm{norm}}>0$ and $j_{\mathrm{norm}}>0$ are normalization parameters for acceleration and jerk, respectively.
The quantity $N_{\mathrm{exit}}(T_{\mathrm{eval}})$ captures the overall throughput, while $N_{\mathrm{col}}(T_{\mathrm{eval}})$ penalizes unsafe behaviors that trigger resets. Overly aggressive policies may increase collision frequency and lead to large accelerations and jerks, which reduces effective progress and decreases the total reward.

\begin{problem}[Multi-Vehicle Intersection Navigation] \label{prob:multi-vehicle-navigation}
Let $\pi^i$ denote the policy of vehicle $i$ and $o^i_k$ its observation at time step $k$. Design a joint control policy $\pi=\{\pi^i\}_{i\in\mathcal{I}}$ with control law
$\bm{u}^i_k=\pi^i(o^i_k)$
that maximizes the total reward $R_{\mathrm{tot}}(T_{\mathrm{eval}})$ defined in \eqref{eq:prob-total-reward} over the evaluation time $T_{\mathrm{eval}}$.
\end{problem}

\section{\ac{cbf}-Informed MARL} \label{sec:cbf-informed-marl}
In \cref{sec:construction-cbf}, we describe how we construct \acp{cbf}, which will later be used to compute safety reward signals that guide the learning process of \ac{marl}, as presented in \cref{sec:cbf-reward}. We add a forward-progress reward and give the final per-agent reward in \cref{sec:forward-progress-reward}.

\subsection{Construction of \acp{cbf}} \label{sec:construction-cbf}
We employ the kinematic bicycle model 
\begin{equation} \label{eq:system-dynamics}
\dot{\bm{x}}=\bm{A}(\bm{x})+\bm{B}\bm{u}
\end{equation}
to model the vehicle dynamics \cite{rajamani2006vehicle}, where $\bm{A}:\mathbb{R}^5\to\mathbb{R}^5$ is given by $\bm{A}(\bm{x})\coloneqq[v\cos(\theta+\beta),v\sin(\theta+\beta),\frac{v}{\ell_{wb}}\tan(\delta)\cos(\beta),0,0]^\top$ and $\bm{B}\coloneqq\big[\bm{0}_{3\times2}^{\top},\bm{I}_2\big]^\top\in\mathbb{R}^{5\times2}$. The state vector is defined as $\bm{x} \coloneqq [x, y, \theta, v, \delta]^\top \in \mathbb{R}^5$, where $x$ and $y$ denote the position, and $\theta$, $v$, and $\delta$ denote the heading, speed, and steering angle, respectively. 
$\bm{u} \coloneqq [u_v, u_\delta]^\top \in \mathbb{R}^2$ denotes the control input vector, with $u_v$ the acceleration and $u_\delta$ the steering rate. $\ell_{wb} >0$ denotes the wheelbase of the vehicle and the slip angle $\beta = \tan^{-1} \left( \tan \delta \cdot \ell_r / \ell_{wb} \right) \in \mathbb{R}$, with $\ell_r >0$ denoting the rear wheelbase. 

A \ac{cbf} is a scalar function $h(\bm x,t)$ that defines a safe set
\begin{equation*}
\mathcal{C}(t)\coloneqq \{\bm x\mid h(\bm x,t)\ge 0\}
\end{equation*}
for a dynamic system of interest. A \ac{cbf} enforces safety by rendering $\mathcal{C}(t)$ forward invariant, i.e., if $\bm x(t_0)\in\mathcal{C}(t_0)$ then $\bm x(t)\in\mathcal{C}(t)$ for all $t\ge t_0$ \cite{ames2019control}.

We construct \acp{cbf} for system \eqref{eq:system-dynamics} and apply our \ac{ttcbf} proposed in \cite{xu2026ttcbf}. Note that our method is generic with respect to \ac{cbf} constructions, i.e., other \ac{cbf} methods such as \cite{xiao2022highorder,nguyen2016exponential} can also be applied. Let $\Delta t$ denote the sampling period. Consider a candidate \ac{cbf} $h$ with relative degree $r$, that is, we need to differentiate $h$ along the system dynamics $r$ times until the control $\bm u$ appears \cite{ames2019control}. We define the associated \ac{cbf} constraint value
\begin{equation}\label{eq:ttcbf}
\psi_h(\bm x,\bm u)\coloneqq \Delta t\,\dot{h}+\cdots+\frac{1}{r!}\Delta t^{r}h^{(r)}+\alpha(h)+R_\mathrm{T},
\end{equation}
where $\alpha(\cdot)$ is an extended class $\mathcal{K}$ function, i.e., a continuous, strictly increasing function defined on an interval containing zero with $\alpha(0)=0$, and $R_\mathrm{T}$ is a Taylor remainder that can be computed as in \cite{xu2026ttcbf}. The \ac{cbf} condition is
\begin{equation} \label{eq:cbf-condition}
\psi_h(\bm x,\bm u)\ge 0.
\end{equation}
Instead of formulating a typical \ac{qp} using \eqref{eq:cbf-condition}, we substitute the \ac{rl} action $\bm u_k$ into \eqref{eq:cbf-condition} to evaluate the corresponding $\psi_h$ values. We then convert these values to reward signals, as detailed in \cref{sec:cbf-reward}.

We now present how we construct \acp{cbf} relevant for solving \cref{prob:multi-vehicle-navigation}. We ignore time dependence to simplify notation. To account for rectangular vehicle footprints and facilitate the construction of \acp{cbf}, we approximate each vehicle occupancy $\mathcal{O}^i(\bm x^i)\subset\mathbb{R}^2$ by a union of $n_\mathrm{cir}$ circles:
\begin{equation*}
\bigcup_{j=1}^{n_{\mathrm{cir}}}\mathcal{O}^{i,j}_{\mathrm{cir}}(\bm x^i)\supseteq \mathcal{O}^i(\bm x^i),
\end{equation*}
where circle $j$ of vehicle $i$ has center $c^{i,j}(\bm x^i)\in\mathbb{R}^2$ and radius $r^{i,j}>0$, following \cite{xu2025realtime}.

\paragraph{Collision avoidance with road boundaries}
For each vehicle $i$, we represent the left and right road boundaries of its drivable corridor $\mathcal{D}^i$ by two polylines
$L^{i,\mathrm{L}}_{\mathrm{road}}$ and $L^{i,\mathrm{R}}_{\mathrm{road}}$ \cite{xu2025realtime}. Let $d_{\mathrm{pseudo}}(p,L)$ denote the pseudo-distance from a point $p\in\mathbb{R}^2$ to a polyline $L$. For each circle $j$ and road boundary side $s\in\{\mathrm{L},\mathrm{R}\}$, we define the \ac{cbf}
\begin{equation*}
h^{i,j,s}_{\mathrm{road}}(\bm x^i)\coloneqq d_{\mathrm{pseudo}}\big(c^{i,j}(\bm x^i),L^{i,s}_{\mathrm{road}}\big)-r^{i,j}
\end{equation*}
and aggregate across circles to obtain one \ac{cbf} per side:
\begin{equation}\label{eq:road_agg}
h^{i,s}_{\mathrm{road}}(\bm x^i)\coloneqq \min_{j\in\{1,\dots,n_{\mathrm{cir}}\}} h^{i,j,s}_{\mathrm{road}}(\bm x^i),\quad s\in\{\mathrm{L},\mathrm{R}\}.
\end{equation}
We denote the corresponding \ac{cbf} constraint values as
\begin{equation}\label{eq:psi_road}
\psi^{i,s}_{\mathrm{road}}(\bm x^i,\bm u^i)\coloneqq \psi_{h^{i,s}_{\mathrm{road}}}(\bm x^i,\bm u^i),\quad s\in\{\mathrm{L},\mathrm{R}\},
\end{equation}
which can be computed by applying \eqref{eq:road_agg} to \eqref{eq:ttcbf}. One can show that this \ac{cbf} has relative degree two. Thus, it suffices to compute $h$'s time derivatives up to second order, i.e., $\dot h$ and $\ddot h$. Given the dynamics in \eqref{eq:system-dynamics}, $\dot h$ and $\ddot h$ follow directly from standard calculus and are omitted for brevity.

\paragraph{Collision avoidance between vehicles}
For two vehicles $i,j\in\mathcal{I}$ with $i\neq j$ and circle indices $a\in\{1,\dots,n_{\mathrm{cir}}\}$ of vehicle $i$ and $b\in\{1,\dots,n_{\mathrm{cir}}\}$ of vehicle $j$, we define
\begin{equation*}
h^{i,j}_{a,b}(\bm x^i,\bm x^j)\coloneqq \|c^{i,a}(\bm x^i)-c^{j,b}(\bm x^j)\|_2-(r^{i,a}+r^{j,b})
\end{equation*}
as \acp{cbf}. We aggregate over circle pairs to obtain one \ac{cbf} per ordered vehicle pair:
\begin{equation}\label{eq:pair_agg}
h^{i,j}_{\mathrm{veh}}(\bm x^i,\bm x^j)\coloneqq \min_{a,b\in\{1,\dots,n_{\mathrm{cir}}\}} h^{i,j}_{a,b}(\bm x^i,\bm x^j).
\end{equation}
Similarly, applying \eqref{eq:pair_agg} to \eqref{eq:ttcbf} yields the corresponding \ac{cbf} constraint value, denoted by
\begin{equation}\label{eq:psi_veh}
\psi^{i,j}_{\mathrm{veh}}(\bm x^i,\bm x^j,\bm u^i, \bm u^j)\coloneqq \psi_{h^{i,j}_{\mathrm{veh}}}(\bm x^i,\bm x^j,\bm u^i, \bm u^j),
\end{equation}
where we emphasize that $\psi^{i,j}_{\mathrm{veh}}$ is evaluated for both vehicles $i$ and $j$ using their joint \ac{rl} actions $\bm u^i$ and $\bm u^j$.

\subsection{\ac{cbf}-Informed Safety Reward} \label{sec:cbf-reward}
This section describes how we compute the safety reward from the \ac{cbf} constraint values in \eqref{eq:psi_road} and \eqref{eq:psi_veh}.

A positive \ac{cbf} constraint value indicates that the \ac{cbf} condition \eqref{eq:cbf-condition} is satisfied, while a negative value indicates a violation. We map each constraint value $\psi$ to a penalty using the linear clipping function
\begin{equation}\label{eq:clip_function}
\rho(\psi)\coloneqq -\min\Big\{\max\Big\{-\frac{\psi}{\psi_\mathrm{cbf}^\mathrm{th}},0\Big\},1\Big\}.
\end{equation}
Here, $\psi_\mathrm{cbf}^\mathrm{th}>0$ is a clipping threshold that sets the violation scale. It normalizes negative constraint values and defines the range over which violations are penalized linearly before saturation. If $\psi\ge 0$, the constraint is satisfied and $\rho(\psi)=0$. If $\psi<0$, then $\rho(\psi)\in[-1,0)$ penalizes the violation magnitude proportionally to $-\psi$, and the penalty saturates at $-1$ once $-\psi\ge\psi_\mathrm{cbf}^\mathrm{th}$.

At time step $k$, we compute the road-boundary \ac{cbf} constraint values $\psi^{i,\mathrm{L}}_{\mathrm{road},k}$ and $\psi^{i,\mathrm{R}}_{\mathrm{road},k}$ for agent $i$ using \eqref{eq:psi_road}, and apply \eqref{eq:clip_function} to obtain
\begin{equation*}
r^{i,\mathrm{L}}_{\mathrm{road},k}\coloneqq \rho\big(\psi^{i,\mathrm{L}}_{\mathrm{road},k}\big),\quad
r^{i,\mathrm{R}}_{\mathrm{road},k}\coloneqq \rho\big(\psi^{i,\mathrm{R}}_{\mathrm{road},k}\big).
\end{equation*}
For collision avoidance with other agents, we compute the pairwise \ac{cbf} constraint values in \eqref{eq:psi_veh} and average the resulting penalties:
\begin{equation*}
r^{i}_{\mathrm{veh},k}\coloneqq \frac{1}{|\mathcal{I}\setminus\{i\}|}\sum_{j\in\mathcal{I}\setminus\{i\}} \rho\big(\psi^{i,j}_{\mathrm{veh},k}\big).
\end{equation*}
We then compute the overall \ac{cbf}-informed reward as
\begin{equation}\label{eq:cbf_reward_total}
r^{i}_{\mathrm{cbf},k}\coloneqq \frac{1}{3}\Big(r^{i}_{\mathrm{veh},k}+r^{i,\mathrm{L}}_{\mathrm{road},k}+r^{i,\mathrm{R}}_{\mathrm{road},k}\Big)
\end{equation}
for agent $i$, which provides action-level feedback on how well its action $\bm u^i$ maintains the joint safety of the multi-agent system.

\subsection{Forward-Progress Reward} \label{sec:forward-progress-reward}
In addition to the safety-related reward $r^{i}_{\mathrm{cbf},k}$, we include a forward-progress reward defined as a weighted sum of projections of the one-step movement onto reference directions:
\begin{equation}\label{eq:reward_progress}
r^{i}_{\mathrm{prog},k}=w_{\mathrm{prog}}\frac{1}{v^{i}_{\max}\Delta t}\sum_{m=1}^{M} w_m s^{i}_{k,m},
\end{equation}
where $s^{i}_{k,m}\coloneqq \langle \Delta p^i_k, \hat d^{i}_{k,m}\rangle$ denotes the projection of the one-step movement $\Delta p^i_k\coloneqq p^i_k-p^i_{k-1}$ onto the normalized reference-direction vector $\hat d^{i}_{k,m}\coloneqq (q^{i}_{k,m}-p^i_{k-1})/\|q^{i}_{k,m}-p^i_{k-1}\|_2$. Here, $p^i_k\in\mathbb{R}^2$ is the position of vehicle $i$ at time step $k$, and $\{q^{i}_{k,m}\in\mathbb{R}^2\}_{m=1}^{M}$ are $M$ short-term reference points sampled from $\mathrm{rp}^i$ ahead of $p^i_{k-1}$. The fixed weight vector $w=[w_1,\ldots,w_M]^\top\in\mathbb{R}^{M}$ assigns larger weights to smaller $m$ to prioritize nearer reference points. $v^{i}_{\max}\Delta t$ is the maximum possible one-step movement of an agent, which acts as a normalizer. 
The learned policy uses these short-term reference points as part of its observation and directly outputs control inputs while learning to avoid other vehicles and road boundaries.
The final reward for agent $i$ at time step $k$ is the sum of the safety-related reward in \eqref{eq:cbf_reward_total} and the forward-progress reward in \eqref{eq:reward_progress}: $r^{i}_{k}\coloneqq r^{i}_{\mathrm{cbf},k}+r^{i}_{\mathrm{prog},k}$.

\section{Evaluation} \label{sec:eva}
In this section, we evaluate the policies learned with our \ac{cbf}-informed reward, denoted as \emph{CBF (our)}, and compare against two heuristic baselines: \emph{Baseline Distance} \cite{wu2023uncertaintyaware} and \emph{Baseline TTC} \cite{zhu2020safe}, which are defined in \cref{sec:eva-baselines-training}. \cref{sec:eva-training} presents the training setup and results. \cref{sec:eva-total-reward} evaluates task-level performance using total reward, \cref{sec:eva-cbf-act} evaluates how strongly each learned policy would rely on a posterior \ac{cbf}-based safety filter, and \cref{sec:eva-representative-behavior} illustrates representative interaction behavior. Code for reproducing the experimental results and a video demonstration are available at \href{https://github.com/bassamlab/SigmaRL}{https://github.com/bassamlab/SigmaRL}.

\subsection{Baseline Methods} \label{sec:eva-baselines-training}
\paragraph{Baseline Distance \cite{wu2023uncertaintyaware}}
\emph{Baseline Distance} penalizes small distances to road boundaries and other agents using a clipped linear shaping function, while keeping the same forward-progress term as in \eqref{eq:reward_progress}. At time step $k$, the reward for agent $i$ is
\begin{equation}\label{eq:rew-distance}
\begin{aligned}
&r^i_k=r^i_{\mathrm{prog},k}+r^i_{\mathrm{dist},k}, \text{ where } \\
&r^i_{\mathrm{dist},k}=\frac{1}{3}\Big(r^{i,\mathrm{L}}_{\mathrm{dist,road},k}+r^{i,\mathrm{R}}_{\mathrm{dist,road},k}+r^i_{\mathrm{dist,veh},k}\Big).   
\end{aligned}
\end{equation}
We define $\rho'(z)\coloneqq -\min\{\max\{z,0\},1\}$.
Let $d^{i,s}_{\mathrm{road},k}\ge 0$ denote the distance from vehicle $i$ to road boundary side $s\in\{\mathrm{L},\mathrm{R}\}$. We compute $r^{i,s}_{\mathrm{dist,road},k}=\rho'\Big(\frac{d_\mathrm{road}^\mathrm{th}-d^{i,s}_{\mathrm{road},k}}{d_\mathrm{road}^\mathrm{th}}\Big)$,
where $d_\mathrm{road}^\mathrm{th}>0$ is the road-distance threshold.
Let $d^{i,j}_{\mathrm{veh},k}\ge 0$ denote the inter-vehicle distance between vehicles $i$ and $j$. We compute
$r^i_{\mathrm{dist,veh},k}=\frac{1}{|\mathcal{I}\setminus\{i\}|}\sum_{j\in\mathcal{I}\setminus\{i\}} \rho'\Big(\frac{d_\mathrm{veh}^\mathrm{th}-d^{i,j}_{\mathrm{veh},k}}{d_\mathrm{veh}^\mathrm{th}}\Big)$,
where $d_\mathrm{veh}^\mathrm{th}>0$ is the vehicle-distance threshold.
Therefore, \emph{Baseline Distance} has two safety-related hyperparameters, $d_\mathrm{road}^\mathrm{th}$ and $d_\mathrm{veh}^\mathrm{th}$.

\begin{table}[t]
    \caption{Simulation parameters.}
    \centering
    \begin{tabular}{ll}
        \toprule
        Parameter & Value \\
        \midrule
        $w_\mathrm{comf}, a_\mathrm{norm}, j_\mathrm{norm}$ & \num{0.2}, \SI{3.0}{\meter\per\second\squared}, \SI{20.0}{\meter\per\second\cubed} \\
        Wheelbase $\ell_{wb}$, rear wheelbase $\ell_r$ & \SI{0.16}{\meter}, \SI{0.08}{\meter} \\
        Max. speed $v_{\max}$ & \SI{1.0}{\meter\per\second} \\
        Max. (min.) acceleration & \SI{5.0}{\meter\per\second\squared} (\SI{-5.0}{\meter\per\second\squared}) \\
        Max. (min.) steering rate & $0.5\pi\,\si{\radian\per\second} \, (-0.5\pi\,\si{\radian\per\second})$\\
        \#short-term reference points $M$ & 3 \\
        Weight $w\coloneqq [w_1,w_2,w_3]^\top$, $w_{\mathrm{prog}}$ & $[\frac{3}{6},\frac{2}{6},\frac{1}{6}]^\top, 0.1$ \\
        $\psi_\mathrm{cbf}^\mathrm{th}$ of \emph{CBF (our)} & $\{0.04,0.06,\ldots,0.20\}$ \\
        $d_\mathrm{veh}^\mathrm{th}$ of \emph{Baseline Distance} & $\{0.02,0.05,0.1,0.15,0.2,0.3\}$ \\
        $t_\mathrm{ttc}^\mathrm{th}$ of \emph{Baseline TTC} & $\{2,3,4,5,6\}$ \\
        $d_\mathrm{road}^\mathrm{th}$ shared by both baselines & $\{0.003,0.005,0.01,0.02\}$ \\
        $T_{\mathrm{eval}}$, sampling period $\Delta t$ & \SI{60}{\second}, \SI{0.1}{\second} \\
        \bottomrule
    \end{tabular}
    \label{tab:parameters}
\end{table}

\paragraph{Baseline TTC \cite{zhu2020safe}}
\emph{Baseline TTC} penalizes imminent collisions using \ac{ttc} under a constant-velocity assumption. At time step $k$, the reward for agent $i$ is
\begin{equation*}
\begin{aligned}
&r^i_k=r^i_{\mathrm{prog},k}+r^i_{\mathrm{ttc},k}, \text{ where } \\
&r^i_{\mathrm{ttc},k}=\frac{1}{3}\Big(r^{i,\mathrm{L}}_{\mathrm{dist,road},k}+r^{i,\mathrm{R}}_{\mathrm{dist,road},k}+r^i_{\mathrm{ttc,veh},k}\Big).   
\end{aligned}
\end{equation*}
It differs from \emph{Baseline Distance} only by replacing the vehicle-collision penalty term (the last term of $r^i_{\mathrm{dist},k}$ in \eqref{eq:rew-distance}) with
$r^i_{\mathrm{ttc,veh},k}=\frac{1}{|\mathcal{I}\setminus\{i\}|}\sum_{j\in\mathcal{I}\setminus\{i\}}\rho'\Big(\frac{t_\mathrm{ttc}^\mathrm{th}-t^{i,j}_{\mathrm{ttc},k}}{t_\mathrm{ttc}^\mathrm{th}}\Big)$,
where $t^{i,j}_{\mathrm{ttc},k}>0$ denotes the \ac{ttc} between vehicles $i$ and $j$ at time step $k$, and $t_\mathrm{ttc}^\mathrm{th}>0$ denotes a threshold. Therefore, \emph{Baseline TTC} has two safety-related hyperparameters, $d_\mathrm{road}^\mathrm{th}$ and $t_\mathrm{ttc}^\mathrm{th}$.

\begin{figure*}[t!]
    \centering
    \begin{subfigure}{0.325\linewidth}
        \centering
        \includegraphics[width=\linewidth]{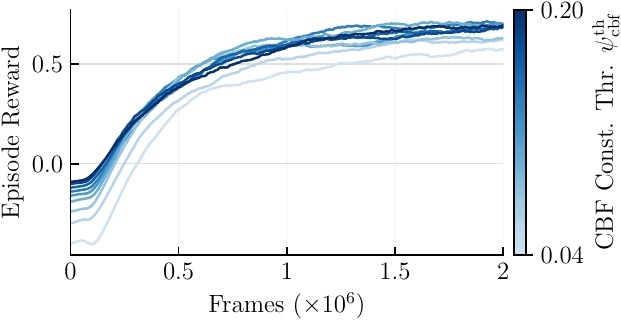}
        \caption{\emph{CBF (our)}.}\label{fig_training_reward_curve_CBF_our}
    \end{subfigure}
    \hfill
    \begin{subfigure}{0.325\linewidth}
        \centering
        \includegraphics[width=\linewidth]{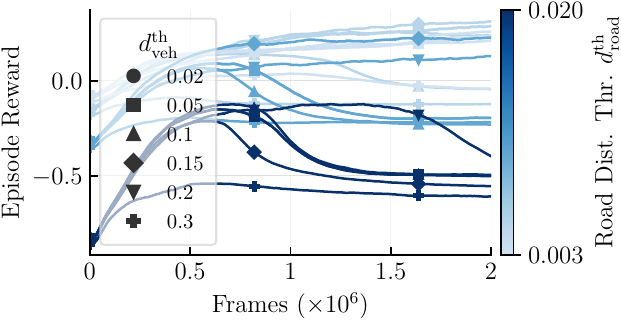}
        \caption{\emph{Baseline Distance}.}\label{fig_training_reward_curve_Distance}
    \end{subfigure}
    \hfill
    \begin{subfigure}{0.325\linewidth}
        \centering
        \includegraphics[width=\linewidth]{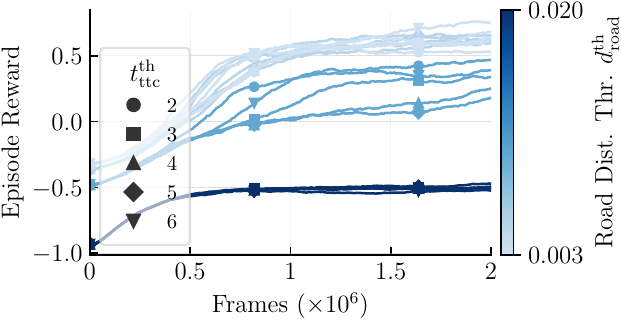}
        \caption{\emph{Baseline TTC}.}\label{fig_training_reward_curve_TTC}
    \end{subfigure}
    \caption{Training reward curves of our method (a) and two baseline methods (b) and (c).}\label{fig_eva_training_curves}
\end{figure*}

\subsection{Training} \label{sec:eva-training}
We train policies in the intersection environment in \cref{fig_intersection} using multi-agent \ac{ppo} with a centralized critic and decentralized actors \cite{lowe2017multiagent}. The observation design follows \cite{xu2024sigmarl}, assuming each \ac{cav} obtains other \acp{cav}' states through communication. All \ac{ppo} parameters are identical across methods, so only the safety-related reward term differs.
\emph{CBF (our)} has one safety-related hyperparameter, $\psi_\mathrm{cbf}^\mathrm{th}$, whereas both baselines have two: $d_\mathrm{road}^\mathrm{th}$ and either $d_\mathrm{veh}^\mathrm{th}$ or $t_\mathrm{ttc}^\mathrm{th}$. \cref{tab:parameters} summarizes the tested ranges and other training and evaluation parameters.

\cref{fig_eva_training_curves} shows the mean episode reward averaged across agents during training. \emph{CBF (our)} exhibits the most stable learning behavior, while the two baselines show larger variability across hyperparameters. This supports our claim that evaluating joint \ac{marl} actions through \ac{cbf} constraint values provides a more consistent safety-related learning signal than distance- or \ac{ttc}-based heuristic proxies.

\begin{figure*}[t!]
    \centering
    \begin{subfigure}{0.325\linewidth}
        \centering
        \includegraphics[width=\linewidth]{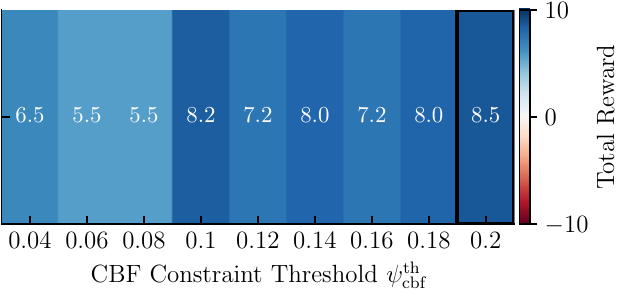}
        \caption{\emph{CBF (our)} ($\psi_\mathrm{cbf}^\mathrm{th}$).}\label{fig_colormap_total_reward_cbf}
    \end{subfigure}
    \hfill
    \begin{subfigure}{0.325\linewidth}
        \centering
        \includegraphics[width=\linewidth]{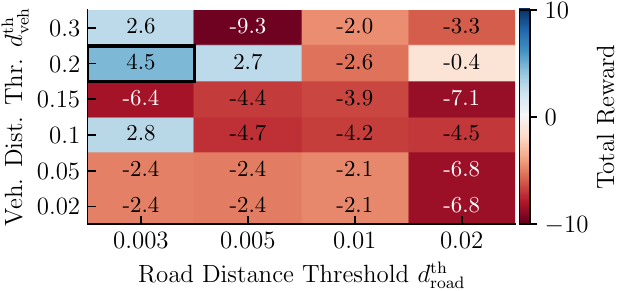}
        \caption{\emph{Baseline Distance} ($d_\mathrm{road}^\mathrm{th}$, $d_\mathrm{veh}^\mathrm{th}$).}\label{fig_colormap_total_reward_distance}
    \end{subfigure}
    \hfill
    \begin{subfigure}{0.325\linewidth}
        \centering
        \includegraphics[width=\linewidth]{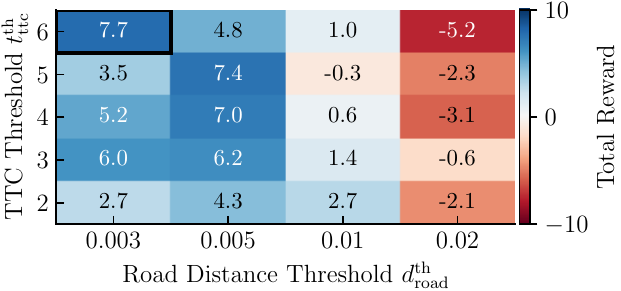}
        \caption{\emph{Baseline TTC} ($d_\mathrm{road}^\mathrm{th}$, $t_\mathrm{ttc}^\mathrm{th}$).}\label{fig_colormap_total_reward_ttc}
    \end{subfigure}
    \caption{Total reward across hyperparameter settings. Best values are marked by black rectangles.}\label{fig_eva_total_reward}
\end{figure*}

\begin{figure*}[t!]
    \centering
    \begin{subfigure}{0.325\linewidth}
        \centering
        \includegraphics[width=\linewidth]{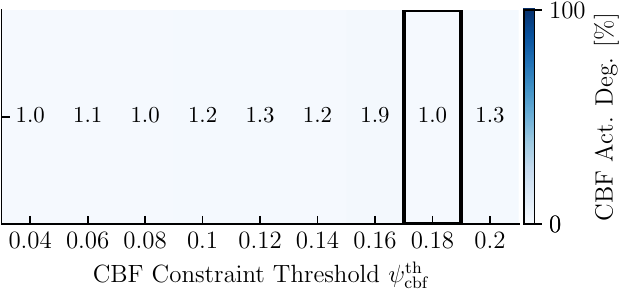}
        \caption{\emph{CBF (our)} ($\psi_\mathrm{cbf}^\mathrm{th}$).}\label{fig_colormap_cbf_activation_degree_cbf}
    \end{subfigure}
    \hfill
    \begin{subfigure}{0.325\linewidth}
        \centering
        \includegraphics[width=\linewidth]{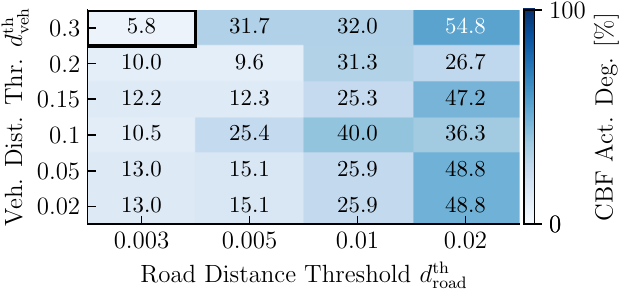}
        \caption{\emph{Baseline Distance} ($d_\mathrm{road}^\mathrm{th}$, $d_\mathrm{veh}^\mathrm{th}$).}\label{fig_colormap_cbf_activation_degree_distance}
    \end{subfigure}
    \hfill
    \begin{subfigure}{0.325\linewidth}
        \centering
        \includegraphics[width=\linewidth]{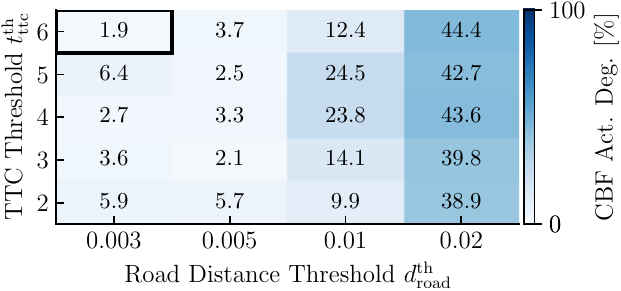}
        \caption{\emph{Baseline TTC} ($d_\mathrm{road}^\mathrm{th}$, $t_\mathrm{ttc}^\mathrm{th}$).}\label{fig_colormap_cbf_activation_degree_ttc}
    \end{subfigure}
    \caption{\ac{cbf} activation degree across hyperparameter settings. Best values are marked by black rectangles.}\label{fig_eva_cbf_act_degree}
\end{figure*}

\subsection{Testing Results: Total Reward} \label{sec:eva-total-reward}
We evaluate task-level performance using the total reward $R_{\mathrm{tot}}(T_{\mathrm{eval}})$ defined in \eqref{eq:prob-total-reward}. For each method, we report (i) performance, defined as the mean total reward averaged over all tested hyperparameter settings, and (ii) sensitivity magnitude, defined as the standard deviation of total reward across tested hyperparameter settings (larger values indicate stronger sensitivity to hyperparameter changes).

\cref{fig_eva_total_reward} depicts the total rewards of our method and the baseline methods. Across the tested hyperparameters, \emph{CBF (our)} achieves the highest mean total reward with \num{7.2}, increasing the mean total reward by \num{9.5} versus \emph{Baseline TTC} with \num{-2.3} and by \num{10.0} versus \emph{Baseline Distance} with \num{-2.8}.
\cref{fig_colormap_total_reward_cbf} further shows that \emph{CBF (our)} maintains positive total reward throughout the tested $\psi_\mathrm{cbf}^\mathrm{th}$ range, while \cref{fig_colormap_total_reward_distance,fig_colormap_total_reward_ttc} show that both baselines include many low-performing and negative-reward regions in their hyperparameter ranges. The best-performing policy of \emph{CBF (our)} is achieved by $\psi_\mathrm{cbf}^\mathrm{th}=\num{0.2}$ with a total reward of \num{8.5} (see the black rectangle in \cref{fig_colormap_total_reward_cbf}).
This outperforms the best-performing policy of \emph{Baseline Distance} (under $d_\mathrm{road}^\mathrm{th}=\num{0.003}$ and $d_\mathrm{veh}^\mathrm{th}=\num{0.2}$), which achieves \num{4.5}, by \SI{88.9}{\percent} and that of \emph{Baseline TTC} (under $d_\mathrm{road}^\mathrm{th}=\num{0.003}$ and $t_\mathrm{ttc}^\mathrm{th}=\num{6}$), which achieves \num{7.7}, by \SI{10.4}{\percent}.

\emph{CBF (our)} also shows much lower sensitivity in the achieved total rewards across all tested hyperparameters. The standard deviation of the total reward is only \num{1.1}. This corresponds to a sensitivity reduction of approximately $66.7\%$ versus \emph{Baseline Distance} with \num{3.3} and $70.3\%$ versus \emph{Baseline TTC} with \num{3.7}, respectively.

\subsection{Testing Results: CBF Activation Degree} \label{sec:eva-cbf-act}
To quantify how strongly a learned policy would rely on a posterior \ac{cbf}-based safety filter, we report the \ac{cbf} activation degree (computed by solving the safety-filter optimization to obtain a correction magnitude but executing the unfiltered action). Lower values indicate that the learned policy produces actions that better align with the \ac{cbf} constraints, which reduces the need for online intervention.

\cref{fig_eva_cbf_act_degree} depicts the resulting \ac{cbf} activation degrees. 
Across the tested hyperparameters, \emph{CBF (our)} clearly yields the lowest activation degree among the compared methods. \cref{fig_colormap_cbf_activation_degree_cbf} shows that the \ac{cbf} activation degree of \emph{CBF (our)} stays in a narrow low range over $\psi_\mathrm{cbf}^\mathrm{th}$. In comparison, \cref{fig_colormap_cbf_activation_degree_distance,fig_colormap_cbf_activation_degree_ttc} show that both baselines often require substantially higher \ac{cbf} activation over large parts of their hyperparameter ranges, indicating stronger dependence on the safety filter. The best-performing policy (in terms of achieved total reward in \cref{fig_colormap_total_reward_cbf}) of \emph{CBF (our)} has a \ac{cbf} activation degree of \SI{1.3}{\percent}, which is \SI{87.0}{\percent} lower than that of \emph{Baseline Distance} (with \SI{10.0}{\percent}) and \SI{31.6}{\percent} lower than that of \emph{Baseline TTC} (with \SI{1.9}{\percent}).

\subsection{Testing Results: Representative Interaction Behavior} \label{sec:eva-representative-behavior}
\begin{figure}[t]
    \centering
    \includegraphics[width=0.78\linewidth]{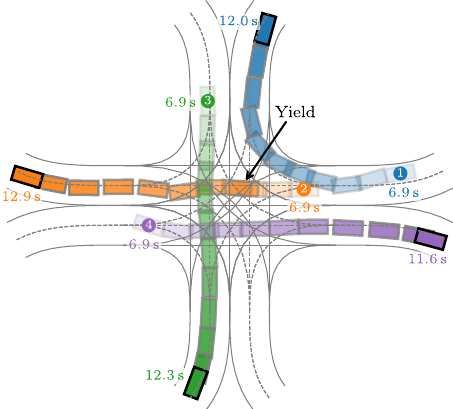}
    \caption{Representative vehicle footprints of \emph{CBF (our)}. Each colored sequence shows accumulated footprints over time, and the start and end times are shown near the first and last footprints, respectively.}
    \label{fig_footprints_rew_method_cbf_k69}
\end{figure}

\cref{fig_footprints_rew_method_cbf_k69} illustrates a representative interaction scenario generated by the policy learned with \emph{CBF (our)}. The accumulated footprints show that vehicle~2 (orange) yields near the conflict region and allows vehicle~3 (green) to pass first. This behavior avoids a potential collision while maintaining progress for the other vehicles, indicating that the learned policy coordinates crossing order rather than only maximizing individual forward motion.

\section{Conclusions}\label{sec:conclusions}
We proposed a Control Barrier Function (CBF)-informed reward design for Multi-Agent Reinforcement Learning (MARL) that converts CBF constraint values under joint MARL actions into reward signals that guide learning toward safety. We evaluated the learned policies in a four-way multi-lane intersection and compared against two heuristic reward baselines (distance-based and time-to-collision-based). Our method achieved the highest task performance: under the best-performing hyperparameter settings, it attained a total reward that is \SI{88.9}{\percent} and \SI{10.4}{\percent} higher than the two baselines, respectively. Moreover, our method reduced reliance on a posterior CBF-based safety filter: it reduced the \ac{cbf} activation degree by \SI{87.0}{\percent} and \SI{31.6}{\percent} relative to the two baselines, respectively. Further, it showed improved robustness to reward hyperparameters, achieving consistently strong performance with lower sensitivity across the tested hyperparameter range.

\bibliographystyle{IEEEtran}
\bibliography{00_literature}

\end{document}